\documentclass[conference]{IEEEtran}
\IEEEoverridecommandlockouts
\usepackage{cite}
\usepackage{amsmath,amssymb,amsfonts}
\usepackage{algorithmic}
\usepackage{graphicx}
\usepackage{textcomp}
\usepackage{xcolor}
\def\BibTeX{{\rm B\kern-.05em{\sc i\kern-.025em b}\kern-.08em
    T\kern-.1667em\lower.7ex\hbox{E}\kern-.125emX}}
\begin{document}

\title{User modeling for social robots\\
\thanks{}
}

\author{\IEEEauthorblockN{ Cristina Gena}
\IEEEauthorblockA{\textit{Computer Science Department} \\
\textit{University of Turin}\\
Italy \\
cristina.gena@unito.it}
\and
\IEEEauthorblockN{Marco Botta}
\IEEEauthorblockA{\textit{Computer Science Department} \\
\textit{University of Turin}\\
Italy \\
marco.botta@unito.it}
\and
\IEEEauthorblockN{Federica Cena}
\IEEEauthorblockA{\textit{Computer Science Department} \\
\textit{University of Turin}\\
Italy \\
federica.cena@unito.it}
\and
\IEEEauthorblockN{Claudio Mattutino}
\IEEEauthorblockA{\textit{Computer Science Department} \\
\textit{University of Turin}\\
Italy \\
claudio.mattutino@unito.it}
}

\maketitle

\begin{abstract}
This paper presents our first attempt to integrate user modeling features in social and affective robots. We propose a cloud-based architecture for modeling the user-robot interaction in order to re-use the approach with different kind of social robots.
\end{abstract}

\begin{IEEEkeywords}
social robot, user modeling, affective computing
\end{IEEEkeywords}

\section{Introduction}
A social robot is an autonomous robot that interact with people by engaging in social-emotive behaviors, skills, capacities, and rules related to its collaborative role \cite{breazeal2004designing}. Thus, social interactive robots need to perceive and understand the richness and complexity of user's natural social behavior, in order to interact with people in a human-like manner \cite{breazeal2004designing}. Detecting and recognizing human action and communication provides a good starting point, but more important is the ability to interpret and react to human behavior, and a key mechanism to carry out these actions is user modeling.
User models are used for a variety of purposes by robots \cite{fong2003survey}.
First, user models can help the robot understand an individual's behavior and dialogue. Secondly, user models are useful for adapting the behavior of the robot to the different abilities, experiences and knowledge of user. Finally, they determine the control form and feedback given to the user (for example, stimulating the interaction).

This paper presents the our attempt to integrate user modeling features in social and affective robots.  We propose a cloud-based architecture for modeling the user-robot interaction in order to re-use the approach with different kind of social robots.

\section{The iRobot application}
\label{project}


The aim of our work is to develop a general purpose cloud-based application,  called iRobot, offering cloud components for managing social, affective and adaptive services for social robots (i.e., Sanbot Elf, Softbak Pepper, etc.). This first version of the application has been developed  and tested for the Sanbot Elf\footnote{http://en.sanbot.com/product/sanbot-elf/} (henceforth simply Sanbot) robot acting as client, thus this client-side application has been called iSanbot. Thanks to the cloud-based service, the robot is  able to handle a basic conversation with users, to recognized them and follow a set of social relationship rules, to detect the user's emotions and modify its behavior according to them. This last task is focused on real-time emotion detection and on how these could change human robot interaction. 
As case study, we started from a scenario wherein the Sanbot robot welcomes people who enter the hall of the Computer Science Dept. of the University of Turin. The robot must be able to recognize the people it has already interacted with and remember past interactions. 

The architecture used for developing the iRobot application is a client-server one. At high level, there are two main components: a client, written in Java for Android, and a server, written in Java, and located on a dedicated machine. The client's goal is to interact with the user and to acquire data aimed at customizing the user-robot interaction, while the server represents the robot's "brain" in the cloud. More in detail, the architecture has been implemented  as follow.
  
The  client side, represented in the current scenario by Sanbot  hosting the iSanbot application, is compatible with any other robot equipped with a network connection and able of acquiring audio and video streaming. The server side, written in Java as well,  acts as a glue between the various software components, each of which creates a different service. In particular, the main services are: the emotion recognition, the face recognition, the user modeling and finally the speech-to-text service. The latter has been created using the Wit service\footnote{https://wit.ai/} (a free Speech to Text service) and communication takes place through a simple HTTP request, sending the audio to be converted into text and waiting for the response. The face recognition, instead, has been developed in Python and implements the librarY "FaceRecognition.py", while the emotion recognition service has been developed in C\# and implements the libraries of Affectiva\footnote{https://www.affectiva.com/}. 
The  application starts with a natural language conversation that has the goal of knowing the user (e.g., face, name, sex, age, interests and predominant mood, etc.) and then recovering all the acquired and elaborated information from the user modeling component for the following interactions, so that the robot will be able to adapt its behavior. 

\subsection{Face recognition service}
First, we implemented the face recognition service. Sanbot, as soon as the user approaches, welcomes and takes a picture of her, and sends it to the Java server, which delegates the Python component to the user's recognition. If the user has not been recognized, a new user will be inserted into the user database. The user is associated to a sets of attributes, such as name, surname, age, sex, profession, interests, moods, etc. to be filled in the following. Some of the attribute indeed may be associated with an explicit question, while other may be implicitly inferred. Thus the client, during the user registration phase, asks the server what the next question is and then it reads it to the user. Once  the user has answered, the audio is sent to the Wit server, converted into text and, one returned, saved into the corresponding attribute associated with the question. For example Sanbot, if does not recognized the user, asks the user her name, then, before taking her a picture, it asks the permission to (\textit{Do you want me to remember you the next time i see you}?), then it asks her profession, and  her favorite color and sport, and stores all these information on the user DB. The detected predominant emotions will be inferred and stored, as well as other biometric data,as described in the following.

On the other hand, if the face recognition component recognizes the user, all her stored information are recovered and the user is welcomed (e.g., \textit{Hello Cristina! Do you feel better today?}) as well the next question to read. When there are no more questions, the conversation ends.

As far as the robot adaptive behavior is concerned, it is now based on simple rules, which  are  used for demonstration purposes  and which  will be then modified  and  extended in our future work. For example, in the current demo, the robot performs a  more informal conversation with young people and a more formal one with older people, it is more playful with young people and more serious with older people. 

For accessibility reasons  and even better understanding, the application replicates all the conversation transcription on the Sanbot tablet screen, both the phrases spoken by Sanbot and those ones acquired by the user.

\subsection{Emotion recognition service}
During conversation with the user, the emotion recognition component is running. This component has been  realized through a continuous streaming of frames that is directed towards the C\# component, which analyzes each frame and extracts the predominant emotions. After that, the server sends a response via a JSON object to the client, which can therefore adapt its behavior and its conversation according to the rule associated to the dominant emotion. For instance, in our application, for demonstration purposes, Sanbot also changes its facial expression with the expression associated with user’s mood.\\
Currently Sanbot is able to recognize six plus one emotions, also known as Ekman's \cite{ekman}  universal emotions and they are: sadness, anger, disgust, joy, fear, surprise and contempt. The predominant emotion of a given interaction will be the one with  a higher certainty factor associated. On the basis of the predominant emotion, returned after the greetings, Sanbot will adapt its behavior accordingly.  

The Affectiva  SDK and API  also provide estimation on gender, age range, ethnicity and a number of other metrics related to the analyzed user. In particular we use the information on gender and age range to implicitly acquire this data about the user and store them on the database. Since also this information is returned with a perceived intensity level, we use  a certainty factor  described in \cite{DBLP:conf/caesar/GenaCMB20}.

\subsection{User modeling} 

Currently, we have created a very simple user model organized as a set of feature-value pairs, and stored in a database. The current stored features are: user name,  age range, gender, favourite sport, favorite color, profession, number of interaction, predominant emotions. Some of the user acquired information are explicitly asked to the user (name, profession, sport, color), other ones (age, sex, emotions) are inferred trough the emotion recognition component described above. Please notice that explicit information as favorite color and sport are used for mere demonstration purposes. \\
In the future we would like to extend the user features with other inferred features such as kind of personality (e.g. Extraversion, Agreeableness, Neuroticism, etc), kind of user dominant emotion, stereotypical user classification  depending on socio-demographic information as sex, age and profession, and so on. Concerning this last point, we will base our stereotypes on shared social studies categorizing somehow the user's interests and preferences on the basis of socio-demographic information. We would also like to import part of the user profile by scanning  social web and social media \cite{socialweb}, with the user's consent.

\section{Conclusion}
\label{conclusion}

Our cloud-based service to model the user and her emotion is just at the beginning. We are now working with the Pepper robot to replicate the same scenario implemented in Sanbot. In the future we would like to replace Affectiva with a similar service we are developing, extend the user model and integrate more sophisticated rules of inferences and adaptation, improve the component of natural language processing.

\bibliographystyle{IEEEtran}

\begin{thebibliography}{1}

\bibitem{breazeal2004designing}
C.~L. Breazeal, \emph{Designing sociable robots}.\hskip 1em plus 0.5em minus
  0.4em\relax MIT press, 2004.

\bibitem{fong2003survey}
T.~Fong, I.~Nourbakhsh, and K.~Dautenhahn, ``A survey of socially interactive
  robots,'' \emph{Robotics and autonomous systems}, vol.~42, no. 3-4, pp.
  143--166, 2003.

\bibitem{ekman}
F.~W. V. . A.~S. Ekman, P., ``Facial signs of
  emotional experience,'' \emph{Journal of
  Personality and Social Psychology}, vol.~39, no.~6, p. 1125--1134, 1980.


\bibitem{DBLP:conf/caesar/GenaCMB20}
C.~Gena, F.~Cena, C.~Mattutino, and M.~Botta, ``Cloud-based user modeling for
  social robots: {A} first attempt (short paper),'' in \emph{Proceedings of the
  Workshop on Adapted intEraction with SociAl Robots, cAESAR 2020, Cagliari,
  Italy, March 17, 2020}, ser. {CEUR} Workshop Proceedings, B.~N.~D. Carolis,
  C.~Gena, A.~Lieto, S.~Rossi, and A.~Sciutti, Eds., vol. 2724.\hskip 1em plus
  0.5em minus 0.4em\relax CEUR-WS.org, 2020, pp. 1--6. [Online]. Available:
  \url{http://ceur-ws.org/Vol-2724/./paper1.pdf}


\bibitem{socialweb}
G.~Sansonetti, F.~Gasparetti, A.~Micarelli, F.~Cena, and C.~Gena, ``Enhancing
  cultural recommendations through social and linked open data,'' \emph{User
  Model. User Adapt. Interact.}, vol.~29, no.~1, pp. 121--159, 2019. [Online].
  Available: \url{https://doi.org/10.1007/s11257-019-09225-8}


\end{thebibliography}

\end{document}